\documentclass[10pt,twocolumn,letterpaper]{article}

\usepackage[pagenumbers]{cvpr} 

\DeclareMathOperator*{\argmin}{arg\,min}

\definecolor{cvprblue}{rgb}{0.21,0.49,0.74}
\usepackage[pagebackref,breaklinks,colorlinks,allcolors=cvprblue]{hyperref}

\title{Gimbal360: Canonicalizing Planar Diffusion for Spherical Panorama Completion}
\author{Yuqin Lu \quad Haofeng Liu \quad Yang Zhou \quad Yihua Dai \quad Guiqing Li \quad Shengfeng He \quad Jun Liang\\
{\tt\small \url{https://orange-3dv-team.github.io/Gimbal360}}
}

\begin{document}

\twocolumn[{%
\renewcommand\twocolumn[1][]{#1}%
\maketitle
\includegraphics[width=.98\linewidth]{fig/teaser.pdf}
\vspace{-1em}
\captionof{figure}{We present \textbf{Gimbal360}, a unified framework for completing a full $360^\circ$ panorama from a single unposed perspective image. Directly applying planar diffusion priors without canonicalization leads to mismatched geometric distortions and visible discontinuities at the panoramic boundary. Gimbal360 resolves these incompatibilities in a canonical spherical representation, producing structurally coherent and seamless panoramic completions. Please zoom in for a better view. \vspace{1.5em}}
\label{fig:teaser}
}]

\begin{abstract}
Diffusion models provide powerful priors for 2D image completion, but these priors are learned on bounded planar images and do not transfer directly to $360^\circ$ panoramas. Perspective observations and spherical panoramas differ in both projective geometry and topology: viewpoint-dependent distortion complicates spatial correspondence, while Equirectangular Projection (ERP) panoramas exhibit intrinsic $S^1$ periodicity that standard Euclidean architectures do not preserve. We present \emph{Gimbal360}, a unified framework that adapts planar diffusion priors to spherical panoramic completion by standardizing these geometric and topological structures. Our Canonical Viewing Space expresses projective distortion as a fixed function of latitude, providing a consistent interface between perspective inputs and spherical panoramas. To map unposed in-the-wild images into this space, Differentiable Projective Canonicalization projects a dense correspondence field onto a 3-DoF rigid projection manifold without requiring camera parameters at inference. We further introduce Topologically Equivariant Generation, which enforces latent shift equivariance to preserve continuity across the periodic ERP boundary. Together, these designs allow diffusion to operate in a representation whose geometry and topology are explicitly aligned with the spherical domain. We also introduce Horizon360, a curated large-scale dataset of gravity-aligned panoramic environments. Extensive experiments show that Gimbal360 achieves state-of-the-art visual fidelity and seam continuity in $360^\circ$ scene completion.
\end{abstract}

\section{Introduction}
\label{sec:intro}
The advent of large-scale diffusion models \citep{rombach2022high} has catalyzed major advances in generative computer vision, particularly in image outpainting and completion \citep{lugmayr2022repaint,zheng2019pluralistic}. Extending these models to $360^\circ$ panoramic completion promises immersive content creation for world generation, virtual reality, and image-based lighting \citep{hunyuanworld2025tencent,worldlabs}, yet remains fundamentally challenging. Because high-quality omnidirectional data remain scarce relative to perspective
imagery, the prevailing paradigm reuses pre-trained planar diffusion priors \citep{rombach2022high,flux2024} for panoramic completion. However, this reuse exposes a fundamental mismatch: such priors are learned on bounded
Euclidean grids ($\mathbb{R}^2$), whereas a panorama lies on the spherical manifold ($S^2$), parameterized by equirectangular projection (ERP). We argue that the recurring failure modes of omnidirectional completion,
including structural collapse, architectural hallucination, and boundary seams, trace back to two unresolved domain gaps.

The first gap arises from \emph{projective variance}. An unposed perspective image is captured under unknown viewing geometry, so its projection onto the ERP
canvas induces distortions that vary with the unknown pose: straight architectural lines become sinusoidal curves whose shape cannot be anticipated by the prior, violating the Manhattan-world regularities it
has internalized. Existing pipelines sidestep this by assuming leveled inputs or ground-truth viewing parameters
\citep{wupanodiffusion,zheng2025panorama,hunyuanworld2025tencent,lu2025matrix3d}, which limits real-world applicability. Learning the alignment implicitly with a spatial transformer \citep{jaderberg2015spatial} is notoriously unstable because the diffusion objective propagates high-frequency, pixel-level gradients that push a free-form warper toward non-rigid, ``jelly-like'' deformations rather than a coherent global alignment.

The second gap stems from \emph{topological severing}. ERP is periodic along the azimuth since its left and right borders depict the same physical location on the $S^1$ boundary. Standard generative architectures, however, assume a bounded Euclidean plane. For example, 
zero-padding in the VAE injects artificial borders that terminate the canvas at the seam, while absolute positional encodings in diffusion transformers treat opposite borders as maximally distant. Both mechanisms
sever the periodic structure of ERP. Prior work mitigates this through inference-time remedies such as circular blending
\citep{feng2023diffusion360} or synchronized multi-view denoising
\citep{lee2023syncdiffusion}. These post-hoc corrections fundamentally behave as spatial
low-pass filters. While they reduce visible seams, they smooth high-frequency details and cannot resolve the underlying topological
mismatch.

In this work, we present \textbf{Gimbal360}, which bridges both gaps through a single
design philosophy: \emph{canonicalization}. Our key observation is that
once a scene is anchored to a gravity-aligned, canonical reference
frame, ERP distortion no longer depends on the unknown viewing geometry
and reduces to a fixed, predictable function of latitude. Gimbal360
therefore canonicalizes at three levels. At the data level, we
curate \emph{Horizon360}, a large-scale collection of gravity-aligned panoramic
environments that establishes a clean canonical structural prior. At the
geometric level, a Differentiable Projective Canonicalization (DPC)
module anchors unposed inputs into the Canonical Viewing Space. It
predicts a dense correspondence field from input pixels to the ERP canvas
and collapses it, through a differentiable rigid solver, onto a 3-DoF
rigid projection manifold. This rigid bottleneck filters noisy diffusion
gradients into globally coherent alignment updates, enabling stable joint
training with the generator while requiring no pose annotations at
inference. At the topological level, we address the Euclidean boundary assumption by enforcing
topological equivariance directly within the generative pipeline. Our
Topologically Equivariant Generation (TEG) combines circular latent
encoding with a Siamese shift-equivariance objective that penalizes inconsistency under azimuthal shifts, teaching the generator
to respect the intrinsic $S^1$ periodic structure of panoramic imagery. This design preserves boundary continuity at the architectural level and
substantially reduces seam artifacts during generation.

In summary, our contributions are threefold:
\begin{itemize}
  \item We identify the geometric and topological domain gaps between planar diffusion priors and the spherical manifold, and introduce a Canonical Viewing Space together with a large-scale gravity-aligned dataset, Horizon360, to establish a unified structural prior.
  \item We propose Differentiable Projective Canonicalization, which regularizes geometric alignment by constraining dense correspondences to a rigid transformation via a differentiable rigid bottleneck, stabilizing joint training with the generator
  without pose annotations at inference.
  \item We enforce topological equivariance in the generative pipeline through circular latent encoding and Siamese shift-equivariant training, enabling diffusion models to respect the intrinsic $S^1$ periodicity of panoramic imagery.
\end{itemize}

\begin{figure*}
    \centering
    \includegraphics[width=.99\linewidth]{fig/framework.pdf}
    \caption{Overview of Gimbal360. Perspective views
are sampled from Horizon360, our gravity-aligned canonical structural
prior (top left). Differentiable Projective Canonicalization (DPC, bottom
left) feeds the perspective input through a geometric encoder to predict a dense
correspondence field, which the rigid solver collapses into a rigid
coordinate grid that warps the view into the Canonical Viewing Space.
Topologically Equivariant Generation (TEG, right) denoises in this space,
where circular latent encoding wraps the frozen VAE via circular-padding and
cropping, and a Siamese consistency loss aligns the predictions of the
base and circularly shifted streams of a shared DiT. Together these
mechanisms teach the generator to respect the periodic $S^1$ boundary
(bottom right), ensuring structurally consistent and seamless panorama completions.}
    \label{fig:framework}
\end{figure*}

\section{Related Work}
\label{sec:relatedwork}

\paragraph{Image Outpainting and Completion.}
Image outpainting extrapolates content beyond the original boundaries while preserving semantic and structural consistency. Early approaches relied on GANs \citep{goodfellow2020generative} to hallucinate missing
regions \citep{wang2019wide,zheng2019pluralistic,zhao2021large}, and diffusion models have since brought substantial gains in fidelity and coherence \citep{lugmayr2022repaint,rombach2022high,esser2024scaling}, with latent diffusion enabling flexible conditioning on image and text prompts. These approaches, however, are inherently designed for planar images on Euclidean grids ($\mathbb{R}^2$); applied directly to omnidirectional
extrapolation, they struggle with the distortions and closed-loop topology of spherical panoramas, producing structural inconsistencies and semantic drift as the field of view expands.

\paragraph{360$^\circ$ Panorama Generation.}
Panoramic generation has progressed from GAN-based systems
\citep{oh2022bips,chen2022text2light,akimoto2022diverse} to diffusion frameworks for text-to-panorama synthesis
\citep{li2023panogen,feng2023diffusion360}, with dual-branch \citep{zhang2024taming} and correspondence-aware \citep{liu2024panofree} designs improving structural consistency. Extending a single narrow field-of-view image to a full panorama is more constrained: most
diffusion-based outpainting pipelines \citep{wupanodiffusion,zheng2025panorama,sun2025spherical,lu2025matrix3d,feng2025dit360,hunyuanworld2025tencent}
project the input onto the ERP canvas using predefined viewing parameters, implicitly assuming gravity-aligned inputs, which limits their
applicability to unconstrained captures \citep{tan2024imagine360}. Closest to our setting, CamFreeDiff \citep{yuan2025camfreediff} removes this
assumption by regressing the input's spherical placement as a one-shot prediction within the diffusion pipeline. In contrast, our DPC module couples a dense correspondence field with a differentiable rigid solver,
a bottleneck that both regularizes the placement estimate and permits stable joint training with the generative objective.

\paragraph{Geometric Adaptations in Omnidirectional Vision.}
Unlike perspective images, ERP panoramas exhibit severe latitude-dependent distortion and require continuity across the periodic boundary. Prior work
adapts to the distortion either by modifying network operators, such as spherical convolutions \citep{coors2018spherenet}, manifold-guided diffusion \citep{sun2025spherical2}, and geometry-aware attention \citep{wu2024spherediffusion}, or by reparameterizing the sphere into cubemaps \citep{kalischek2025cubediff,huang2025dreamcube} and tangent planes \citep{ccapuk2025tandit}. Boundary continuity is likewise addressed through cylindrical convolutions \citep{liao2023cylin} or inference-time
corrections such as circular blending \citep{feng2023diffusion360}, synchronized denoising \citep{lee2023syncdiffusion}, and circular padding
\citep{wang2024360dvd}, which reduce visible seams but act as post-hoc low-pass corrections. Ensuring strict topological equivariance within the generative modeling process itself remains an open challenge.
\section{Gimbal360}
\label{sec:method}

Given an unposed, narrow field-of-view (NFoV) perspective image $I_p \in \mathbb{R}^{H_p \times W_p \times 3}$, our goal is to synthesize a structurally coherent $360^\circ$ ERP panorama $I_e \in \mathbb{R}^{H_e \times W_e \times 3}$. This task is fundamentally ill-posed: projecting an unposed $I_p$ onto the spherical $S^2$ manifold induces severe non-linear distortions that violate the Euclidean ($\mathbb{R}^2$) structural priors of pre-trained diffusion models. To bridge planar diffusion priors and spherical panoramas, we propose Gimbal360. Instead of forcing the network to implicitly resolve arbitrary projective mappings, we introduce a Canonical Viewing Space, a gravity-aligned spherical coordinate system supported by our newly curated Horizon360 dataset, and resolve the domain gaps through two mechanisms.
A Differentiable Projective Canonicalization (DPC) module
recovers the spherical placement of the input and anchors its features into the canonical space through a rigid bottleneck, and a Topologically Equivariant Generation (TEG) scheme preserves the periodic boundary of the panorama during diffusion. Fig.~\ref{fig:framework} provides an overview.

\subsection{Preliminaries}
\label{sec:preliminaries}
Latent Diffusion Models (LDMs) \citep{rombach2022high} operate in the
compressed latent space of a pre-trained autoencoder. An encoder
$\mathcal{E}$ maps an image $x$ to a spatial latent
$z_0 = \mathcal{E}(x)$, and a decoder $\mathcal{D}$ reconstructs
$x \approx \mathcal{D}(z_0)$. During training, a forward process corrupts
$z_0$ with Gaussian noise over $T$ timesteps to produce $z_t$, and a
conditional denoiser $\epsilon_\theta$ is trained to predict the injected
noise
\begin{equation}
  \mathcal{L}_{\mathrm{LDM}} =
  \mathbb{E}_{z_0,\,\epsilon \sim \mathcal{N}(0,I),\,t}
  \big[\, \lVert \epsilon - \epsilon_\theta(z_t, t, \tau(y)) \rVert_2^2 \,\big],
\end{equation}
where $\tau(y)$ denotes conditioning embeddings, such as text prompts extracted via a CLIP text encoder~\cite{radford2021learning}.

For completion tasks, generation is further conditioned on known visual
context through a binary mask $M$, in which $1$ marks given source pixels
and $0$ the region to synthesize. The reference latent
$z_{\mathrm{ref}} = z_0 \odot M$ is concatenated with $z_t$ and $M$ along
the channel dimension, yielding
$\hat{\epsilon} = \epsilon_\theta(z_t \oplus M \oplus z_{\mathrm{ref}},
t, \tau(y))$. While effective for planar outpainting, this formulation
does not account for the spherical geometry and periodic topology of
omnidirectional imagery, which motivates our canonicalized framework.

\subsection{The Omnidirectional Domain Gaps}
\label{sec:domain_gaps}

To successfully integrate Euclidean ($\mathbb{R}^2$) generative priors with the spherical ($S^2$) omnidirectional manifold, our framework is dedicated to resolving two fundamental geometric discrepancies inherent to $360^\circ$ generation.

\paragraph{Projective Variance.}
The mapping from a perspective image to the ERP canvas is determined by where the observation sits on the sphere. For in-the-wild inputs this spherical placement is unknown, and each placement induces a different
distortion pattern in which straight world lines become sinusoidal curves.
Such input-dependent distortion destabilizes the latent conditioning of LDMs, which implicitly rely on Manhattan-world regularities learned from perspective imagery. We resolve this by reformulating the generation
target around a Canonical Viewing Space $\mathcal{C}$, a gravity-aligned frame in which the environmental vanishing line coincides with the ERP equator. Anchored to $\mathcal{C}$, the distortion no longer depends on
the input placement and reduces to a fixed, predictable function of latitude. We thus seek a proxy transformation $\mathcal{T}_{\mathrm{align}}$ that warps the input features into the canonical space, producing the canonical generative anchor
\begin{equation}
  z_{\mathrm{canonical}} = \mathcal{T}_{\mathrm{align}}(z_{\mathrm{ref}}).
\end{equation}

\paragraph{Topological Severing.}
The ERP format inherently spans the azimuthal angle $\theta \in [-\pi, \pi]$, so its
left and right boundaries depict physically adjacent content on the
periodic $S^1$ boundary. Standard latent generative pipelines break this structure.
Zero-padding during latent compression injects artificial borders, and absolute positional encodings in diffusion transformers treat opposite boundaries as maximally distant, which mathematically severs the $S^1$ periodic boundary and manifests as visible seams in the
generated output. To mitigate this, our generative process is governed by strict Topological Equivariance, ensuring the network natively respects the continuous topology of the manifold.

\subsection{Differentiable Projective Canonicalization}
\label{sec:dpc}

As established, our framework requires a proxy transformation $\mathcal{T}_{\mathrm{align}}$ that anchors the unposed input to the Canonical
Viewing Space $\mathcal{C}$. A natural choice is to train an unconstrained Spatial Transformer Network~\cite{jaderberg2015spatial} end-to-end with the latent diffusion loss. However, $\mathcal{L}_{\mathrm{LDM}}$ is dominated by high-frequency, pixel-level variations, and backpropagating such noisy
gradients into a free-form warper induces non-rigid, ``jelly-like'' deformations rather than a coherent global alignment. Our Differentiable
Projective Canonicalization (DPC) module avoids this failure mode by decoupling dense correspondence prediction from rigid geometric fitting.

\paragraph{Dense Correspondence Prediction.}
Given the perspective input $I_p$, a SegNeXt~\cite{guo2022segnext} encoder
with a lightweight decoder predicts a dense correspondence field
$\mathcal{P}_{\mathrm{dense}} \in \mathbb{R}^{H_p \times W_p \times 2}$ that
assigns to every input pixel its target coordinates on the canonical ERP
canvas. Instead of regressing coordinates from scratch, the network predicts
a bounded residual on top of the base grid of a canonical upright camera,
and a ray-direction positional encoding is injected into the decoder so that
the prediction remains aware of the underlying viewing geometry.

\paragraph{Differentiable Rigid Solver.}
Rather than warping features directly with $\mathcal{P}_{\mathrm{dense}}$,
we collapse it onto the 3-DoF manifold of rigid spherical cameras. We
parameterize the camera as $\phi = (\theta_v, \theta_p, \theta_r)$, whose forward projection
$f(\phi)$ maps perspective pixels onto the ERP canvas, and fit $\phi$ to the
predicted field by nonlinear least squares:
\begin{equation}
  \hat{\phi} \;=\; \argmin_{\phi}\;
  \big\lVert f(\phi) - \mathcal{P}_{\mathrm{dense}} \big\rVert_2^2 ,
  \label{eq:rigid_fit}
\end{equation}
solved with a closed-form Procrustes initialization followed by a few
Levenberg--Marquardt iterations. From $\hat{\phi}$ we analytically
reconstruct a \emph{rigid coordinate grid}, which warps the input latent
into $\mathcal{C}$ to form the canonical generative anchor:
\begin{equation}
  z_{\mathrm{canonical}} \;=\; \mathcal{W}(z_{\mathrm{ref}},\, \hat{\phi}),
\end{equation}
where $\mathcal{W}$ denotes differentiable grid sampling with the rigid
coordinate grid.

\paragraph{Rigid Bottleneck as a Gradient Filter.}
By forcing the dense correspondence field to pass through the rigid
bottleneck, the solver's Jacobian acts as a structural gradient filter
during backpropagation. Because the rigid camera model has only three
degrees of freedom, backpropagated gradients are projected onto the rank-3
subspace spanned by its Jacobian. Thus, noisy, spatially uncorrelated components
of the diffusion gradient are suppressed, and only globally coherent, rigid
alignment updates reach the dense predictor. This design prevents localized
``jelly'' warps while still allowing DPC to learn, jointly with the
generator, the anchoring that best serves panorama completion.

\subsection{Topologically Equivariant Generation}
\label{sec:teg}

Having anchored the input features into $\mathcal{C}$, the remaining challenge is to synthesize the panorama without severing its periodic boundary. We enforce topological equivariance across both the compression
and generation stages.

\paragraph{Circular Latent Encoding.}
We remove the artificial boundary at the compression stage without modifying the pre-trained VAE. Before encoding, the input is circularly pre-padded along the azimuthal axis, and the resulting latent is cropped
back to its nominal extent afterwards. The convolutional receptive fields therefore wrap around the boundary while the frozen autoencoder weights and functionality remain untouched.

\paragraph{Siamese Shift-Equivariant Generation.}
Circular encoding provides the structural pathway for continuous feature propagation, but the generator must also be penalized for boundary-inconsistent predictions. Let $\mathrm{Roll}_\delta(\cdot)$ denote a circular shift along the azimuthal axis by a random offset $\delta$. During training we perform two parallel prediction passes,
\begin{align}
  \hat{\epsilon}_{\mathrm{base}} &=
  \epsilon_\theta(z_t \oplus M \oplus z_{\mathrm{canonical}},\, t,\, \tau(y)), \\
  \hat{\epsilon}_{\mathrm{shifted}} &=
  \epsilon_\theta(\mathrm{Roll}_\delta(z_t \oplus M \oplus
  z_{\mathrm{canonical}}),\, t,\, \tau(y)).
\end{align}
If the network models the underlying $S^1$ topology rather than overfitting to the Euclidean frame, shifting the base prediction by $\delta$ should agree with predicting from the shifted inputs. We penalize the discrepancy with a Siamese consistency loss
\begin{equation}
  \mathcal{L}_{\mathrm{shift}} =
  \mathbb{E}_{z_0,\epsilon,t,\delta}
  \big[\, \lVert \mathrm{Roll}_\delta(\hat{\epsilon}_{\mathrm{base}})
  - \hat{\epsilon}_{\mathrm{shifted}} \rVert_2^2 \,\big].
\end{equation}
This objective enforces shift consistency in the latent space during denoising. By explicitly penalizing boundary-inconsistent generative logic, this formulation intrinsically binds the network to the continuous $S^1$ manifold. In practice, the learned
consistency yields perceptually seamless outputs, which we quantify with seam-aware metrics in our experiments.

\subsection{Overall Training Objective}
\label{sec:objective}

The rigid bottleneck delivers a deliberately low-dimensional gradient to the dense predictor, which regularizes alignment but is insufficient as a sole training signal. We therefore anchor the dense correspondence field with direct supervision against the
ground-truth correspondence $\mathcal{P}_{GT}$, analytically derived from the known sampling geometry of our canonicalized dataset,
\begin{equation}
  \mathcal{L}_{\mathrm{corr}} =
  \mathrm{SmoothL1}(\mathcal{P}_{\mathrm{dense}},\, \mathcal{P}_{GT}).
\end{equation}
The overall objective jointly minimizes the latent diffusion loss, the topological regularizer, and the geometric anchor,
\begin{equation}
  \mathcal{L}_{\mathrm{total}} =
  \mathcal{L}_{\mathrm{LDM}}
  + \lambda_{\mathrm{shift}} \mathcal{L}_{\mathrm{shift}}
  + \lambda_{\mathrm{corr}} \mathcal{L}_{\mathrm{corr}},
\end{equation}
where $\lambda_{\mathrm{shift}}$ and $\lambda_{\mathrm{corr}}$ are scalar
weights. Dense supervision provides the primary high-resolution signal for the correspondence field, while diffusion gradients arriving through the rigid bottleneck refine the anchoring toward placements that best serve
completion, rather than accuracy alone.

\section{The Horizon360 Dataset}
\label{sec:dataset}

Horizon360 constitutes the data level of our canonicalization. The structural prior that Gimbal360 learns is only as reliable as the canonical properties of its training panoramas, yet existing omnidirectional datasets \citep{xiao2012recognizing,chang2017matterport3d}
frequently contain in-the-wild tilts and inconsistent horizon lines. Training a generative prior on such unrectified data contradicts the very constraints the Canonical Viewing Space is meant to provide. We therefore
curate Horizon360, a large-scale panoramic dataset in which every environment satisfies the canonical constraints of $\mathcal{C}$.

\paragraph{Curation and Canonicalization.}
Horizon360 comprises 20k high-fidelity indoor and outdoor panoramic environments sourced from established open datasets \citep{chang2017matterport3d,feng2025dit360,kocabas2021spec,yang2025layerpano3d}, web collections \citep{polyhaven}, and custom renders produced with Unreal
Engine. Building on foundational omnidirectional geometric principles \citep{zhang2014panocontext,zou2018layoutnet}, we adapt vanishing-point and zenith alignment strategies
\citep{sun2019horizonnet,sun2021hohonet,jiang2022lgt} to rectify every raw panorama, rotating the sphere so that the environmental vanishing line coincides with the ERP equator and physical plumb lines remain vertical.
The resulting collection is gravity-aligned by construction. 

\paragraph{Realistic Perspective Sampling.}
To train DPC and the canonicalized generator, we deploy a mixed sampling strategy that improves real-world robustness over standard uniform cropping \citep{jin2023perspective,veicht2024geocalib,tirado2025anycalib}.
Unlike prior works that fix the horizontal field of view, we drive sampling by the vertical field of view $\theta_v$ and aspect ratio $r$, simulating diverse capture devices from wide-angle to telephoto through
the effective focal lengths
\begin{equation}
  f_y = \frac{1}{\tan(\theta_v / 2)}, \qquad f_x = \frac{f_y}{r}.
\end{equation}
Uniform sampling over-represents implausible extreme placements, while purely Gaussian sampling leaves the network brittle to outliers. We therefore draw the sampling parameters
$\Theta = \{\theta_{\mathrm{pitch}}, \theta_{\mathrm{roll}}, \theta_v, r\}$
from a hybrid prior
\begin{equation}
  P(\Theta) = \lambda \cdot \mathcal{N}(\mu_{\mathrm{real}},
  \sigma^2_{\mathrm{real}})
  + (1 - \lambda) \cdot \mathcal{U}(\Theta_{\min}, \Theta_{\max}),
\end{equation}
where $\lambda = 0.7$ focuses training on common, near-upright captures while retaining a $30\%$ exposure to extreme placements, improving robustness to unconstrained user inputs.

\paragraph{Yaw-Centered Canonicalization.}
Finally, we decouple scene content from absolute heading. A circular permutation aligns the sampled view with the canonical center,
\begin{equation}
  I_{\mathrm{centered}} = \mathrm{Roll}(I_{\mathrm{ERP}},\, -\psi_{\mathrm{input}}),
\end{equation}
so the model always receives conditioning at the center of the canvas and the wraparound boundary falls $180^\circ$ behind the primary view. This simplification reduces boundary artifacts and stabilizes the topologically equivariant training of the generator.

\section{Experiments}
\label{sec:exp}

\begin{table*}[t]
  \centering
  \setlength{\tabcolsep}{2.2pt}
  \begin{tabular}{l cccccc cccccc}
    \toprule
    & \multicolumn{6}{c}{Indoor, 5K}
    & \multicolumn{6}{c}{Outdoor, 5K} \\
    \cmidrule(lr){2-7} \cmidrule(lr){8-13}
    Method
    & FID$\downarrow$ & KID($\times 10^2$)$\downarrow$ & O-FID$\downarrow$ & FAED$\downarrow$ & CS$\uparrow$ & DS$\downarrow$
    & FID$\downarrow$ & KID($\times 10^2$)$\downarrow$ & O-FID$\downarrow$ & FAED$\downarrow$ & CS$\uparrow$ & DS$\downarrow$ \\
    \midrule
    HunyuanWorld~1.0
    & 29.23 & 2.1981 & 53.26 & 4.05 & 26.48 & 0.92
    & 41.37 & 2.2218 & 69.44 & 7.07 & 29.13 & 1.00 \\
    WorldGen
    & 27.43 & 1.9683 & 50.50 & 2.45 & 26.44 & 1.32
    & 39.78 & 2.0616 & 67.47 & 6.47 & 29.10 & 1.01 \\
    DiT360
    & 38.43 & 3.4848 & 70.61 & 11.30 & \textbf{31.09} & 3.76
    & 55.17 & 3.7295 & 65.56 & 8.35 & \textbf{31.40} & 2.08 \\
    \midrule
    \textbf{Ours}
    & \textbf{15.92} & \textbf{1.0249} & \textbf{30.10} & \textbf{1.36} & 30.68 & \textbf{0.49}
    & \textbf{23.78} & \textbf{0.9228} & \textbf{47.48} & \textbf{1.75} & 31.29 & \textbf{0.66} \\
    \bottomrule
  \end{tabular}
  \caption{Quantitative comparison on both indoor and outdoor scenes, with 5,000 held-out samples per domain. KID is scaled by $10^2$. Bold marks the best result.}
  \label{tab:quantitative}
\end{table*}

\begin{figure*}[t]
  \centering
  \includegraphics[width=\linewidth]{fig/comparisons.pdf}
  \caption{Qualitative comparison on indoor (left) and outdoor (right) scenes. Baseline completions exhibit warped
structures, mismatched distortion profiles, and boundary discontinuities, whereas ours completes structurally coherent, boundary-continuous panoramas. Please zoom in for details.}
\label{fig:qualitative}
\end{figure*}

\subsection{Experimental Settings}
\paragraph{Implementation Details.}
Our model is built upon Flux.1-fill-dev~\cite{flux2024} and fine-tuned
with LoRA~\cite{hu2022lora} on 8 GPUs for 10,000 steps with a batch size
of 8, using AdamW with a constant learning rate of $1\times10^{-4}$. The
training data comprises 20,000 panoramas from Horizon360, each producing
three perspective samples under the hybrid prior of realistic sampling.
We set the topological weight to $\lambda_{\mathrm{shift}} = 0.5$ and the
geometric weight to $\lambda_{\mathrm{corr}} = 0.1$, balancing the
auxiliary objectives against the diffusion loss.

\paragraph{Evaluation Protocol.}
For evaluation, we construct two held-out test sets: 5000 indoor images randomly sampled from Structured3D~\cite{Structured3D} and 5000 outdoor images from CVRG-Pano~\cite{orhan2021semantic} and LayerPano3D~\cite{yang2025layerpano3d}, totaling 10,000 test images for comprehensive benchmarking. We measure visual quality with FID~\cite{heusel2017gans}, KID~\cite{binkowski2018demystifying}, and FAED~\cite{zhang2024taming}, and
text alignment with CLIPScore~\cite{hessel2021clipscore}. To account for the spherical geometry and boundary continuity that these planar metrics fail to capture, we additionally report OmniFID (O-FID), which extends FID to the spherical domain through cubemap reprojection \citep{christensen2024geometry}, and
the Discontinuity Score (DS), which quantifies
discontinuity across the periodic seam.

\subsection{Main Results}
We compare our method against HunyuanWorld~1.0 \citep{hunyuanworld2025tencent},
WorldGen \citep{worldgen2025ziyangxie}, and DiT360 \citep{feng2025dit360},
the strongest open-source methods built on the same Flux-fill backbone,
which ensures that differences reflect the framework rather than the base
model.

\paragraph{Quantitative Comparison.}
Tab.~\ref{tab:quantitative} reports results on both indoor and outdoor scenes. 
Our method achieves the best scores on all fidelity and geometry metrics, reducing
FID over the strongest baseline WorldGen by $42\%$ indoors and $40\%$ outdoors, with comparable margins on KID, O-FID, and FAED. On the seam-specific DS, our score of 0.49 indoors nearly halves the best competing result, indicating that the improvements
stem from structural and topological consistency rather than texture
fidelity alone. CLIPScore varies only marginally across methods. DiT360
scores highest by a small margin yet exhibits the weakest
structural fidelity and seam continuity, suggesting that text adherence
alone does not translate into coherent panoramas.

\begin{figure}[h]
  \centering
  \includegraphics[trim={0 0 0 0},clip,width=\linewidth]{fig/generalization.pdf}
  \caption{Qualitative results of in-the-wild generalization. Inputs include real-world captures and styled photos. Competing methods produce mismatched surroundings or warped layouts, whereas Gimbal360 completes structure- and style-consistent panoramas. Please zoom in for details.}
  \label{fig:ood}
\end{figure}

\paragraph{Qualitative Comparison}
Fig.~\ref{fig:qualitative} compares panoramic completions on indoor and outdoor scenes, where dashed outlines mark corresponding regions across methods and are
magnified in the side insets. Gimbal360 anchors each unposed input at its
canonical frame, so the completed environments exhibit the
correct latitude-dependent distortion profile, straight structural lines,
and coherent global layout. HunyuanWorld projects inputs with fixed
viewing parameters and thus inherits a mismatched distortion profile,
bending structures around the conditioning region. WorldGen reduces these
errors yet retains local structural drift and occasional discontinuities
at the periodic boundary, and DiT360 fails to preserve the input region and exhibits severe structural inconsistency. The magnified crops further show that our canonicalized completions preserve structural regularity and remain coherent in both scene content and layout. Moreover, our method surpasses all competing baselines in synthesizing continuous content across the seam, without resorting to post-hoc blending.

\begin{table}[t]
  \centering
  \small
  \setlength{\tabcolsep}{1.2pt}
  \begin{tabular}{l cccc cccc}
    \toprule
    & \multicolumn{4}{c}{Indoor, 0.5K} & \multicolumn{4}{c}{Outdoor, 0.5K} \\
    \cmidrule(lr){2-5} \cmidrule(lr){6-9}
    Method 
    & FID$\downarrow$ &KID $\downarrow$ & O-FID$\downarrow$ & DS$\downarrow$
    & FID$\downarrow$ &KID $\downarrow$ & O-FID$\downarrow$ & DS$\downarrow$ \\
    \midrule
    w/o DPC    & 64.48 &1.2716 & 93.03  & 1.63 & 69.82 & 1.9039 & 108.00 & 1.09 \\
    
    Geocalib    & 70.50 & 1.3122 & 100.27 & 1.10 & 78.93 & 2.1227 & 112.35 & 0.98 \\
    PF        & 72.54 & 1.3640 & 102.36 & 0.99 & 81.25 & 2.5321 & 116.56 & 0.94 \\
    w/o TEG    & 55.05 & 0.8106 & 79.49  & 1.75 & \textbf{62.74} &\textbf{0.8194} & 106.62 & 1.49 \\
    \midrule
    \textbf{Ours} 
                          & \textbf{54.18} &\textbf{0.8012} & \textbf{79.45} & \textbf{0.51}
                          & 63.28 & 0.8236 & \textbf{99.04} & \textbf{0.69} \\
    \bottomrule
  \end{tabular}
  \caption{Ablation study and anchoring baselines. Geocalib and PF denote replacing DPC with GeoCalib~\cite{veicht2024geocalib} and
  PerspectiveFields~\cite{jin2023perspective}, respectively, while keeping TEG fixed. Bold marks the best result. KID is scaled by $10^2$.}
  \label{tab:ablation}
\end{table}

\paragraph{In-the-Wild Generalization.}
Fig.~\ref{fig:ood} evaluates zero-shot robustness on in-the-wild inputs outside the training distribution, spanning real indoor photographs with unknown
viewing geometry and stylized illustrations with various aspect ratios. Gimbal360 anchors each input in the canonical viewing space to better leverage the internalized generative priors. Thus, the completions remain faithful to the conditioning in structure, semantics, and visual style, extending the illustrated scene with a plausible global layout. Competing methods instead surround the input with content of mismatched scale or semantics, wash out the scene identity, or warp the global layout, as highlighted by the dashed regions and magnified crops.
These support the practical utility of our canonicalized completion for unconstrained inputs.

\begin{figure}[h]
  \centering
  \includegraphics[trim={0 0 0 0},clip,width=.99\linewidth]{fig/ablation.pdf}
  \caption{Qualitative ablation results. Removing
DPC yields severe structural distortion while removing TEG breaks continuity near the periodic boundary. Our full model produces structurally coherent and
boundary-continuous completions.}
  \label{fig:ablation}
\end{figure}

\subsection{Ablation Studies}

We ablate on a subset of 500 images per domain sampled from our test
sets, comparing five variants under identical settings. \emph{w/o DPC} removes the geometric module, projecting the input with fixed viewing parameters and forcing the generator to resolve projective variance implicitly. \emph{w/o TEG} keeps DPC but trains with standard latent encoding and no shift-equivariance objective. \emph{Geocalib} and \emph{PF} replace DPC with the external estimators
GeoCalib \citep{veicht2024geocalib} and PerspectiveFields \citep{jin2023perspective}, whose predicted parameters drive the projection while TEG remains active. \emph{Ours} is the full model. Quantitative and qualitative results are reported in Tab.~\ref{tab:ablation} and Fig.~\ref{fig:ablation}.

\paragraph{Effect of DPC.}
Removing DPC degrades every metric, with indoor FID rising from 54.18 to
64.48 and outdoor O-FID from 99.04 to 108.00, and yields severe
structural distortion. More importantly, both Geocalib and PF recover plausible placements, yet completion quality drops markedly, with indoor FID rising to 70.50 and 72.54 respectively. This indicates that a geometrically accurate placement is not necessarily the one that best
serves completion. Trained jointly with the generator, DPC learns the
anchoring that minimizes downstream synthesis error, which makes it an
integral component rather than a swappable preprocessing step.

\paragraph{Effect of TEG.}
Removing TEG leaves FID nearly unchanged, and the ablated variant even
attains a marginally lower outdoor FID. This is expected since FID is insensitive to the periodic boundary. The seam-specific DS reveals the actual contribution, dropping from 1.75 to 0.51 indoors and from 1.49 to 0.69 outdoors once TEG is enabled. These gains indicate that circular latent encoding and the Siamese objective enforce a continuity that post-hoc corrections cannot achieve, and that the two mechanisms complement each other, with the full model achieving the best overall results.

\section{Conclusion}
We presented Gimbal360, a framework for completing $360^\circ$ panoramas from a single unposed perspective image by explicitly addressing the geometric and topological mismatch between Euclidean diffusion priors and spherical panoramas. By introducing a Canonical Viewing Space, a Differentiable Projective Canonicalization module for anchoring inputs, and a topologically equivariant generation strategy that preserves the intrinsic $S^1$ periodicity, our method produces structurally consistent and seamless omnidirectional completions.


{
    \small
    \bibliographystyle{ieeenat_fullname}
    \bibliography{main}
}


\clearpage
\appendix
\maketitlesupplementary

This supplementary document provides extended technical details,
comprehensive qualitative evaluations, and dataset curation
methodologies. Sec.~\ref{sec:supp_exp_setup} provides both training and inference details. Sec.~\ref{sec:supp_dataset} elaborates on the
curation, canonicalization, and captioning of the Horizon360 dataset.
Sec.~\ref{sec:supp_proof} presents an analysis of our Siamese
shift-equivariance objective on the $S^1$ boundary, followed by our
efficient positional-encoding-based implementation. Sec.~\ref{sec:supp_qualitative}
presents additional qualitative results and a discussion of limitations.

\section{Extended Experimental Setups}
\label{sec:supp_exp_setup}

\subsection{Training Details}
To fine-tune Gimbal360 while managing the memory footprint of the
Flux.1-fill-dev backbone \citep{flux2024}, we apply LoRA \citep{hu2022lora}
to the attention matrices (query, key, value) and feed-forward networks of
the frozen transformer blocks, with rank $r=64$ and scaling $\alpha=64$.
Target panoramas are processed at an ERP resolution of $960 \times 1920$.

The pre-trained VAE remains entirely frozen. We do not modify its layers
or weights. Continuity across the $S^1$ boundary is instead obtained by
circularly pre-padding the input along the azimuthal axis before encoding
and cropping the latent back to its nominal extent afterwards, so the
convolutional receptive fields wrap around the boundary while the
pre-trained latent distribution is preserved.

The overall objective jointly minimizes the diffusion loss, the
topological Siamese regularizer, and the geometric correspondence anchor,
with $\lambda_{\mathrm{shift}} = 0.5$ and
$\lambda_{\mathrm{corr}} = 0.1$ balancing the auxiliary gradients against
the diffusion loss. Training runs on 4 NVIDIA H20 GPUs with
PyTorch 2.7.0. 

\subsection{Inference Details}
We use the Euler sampler for 30 denoising steps with a classifier-free
guidance scale of 30.0, the default configuration of Flux-fill. To
exploit the learned shift equivariance, we apply a random circular shift
$\delta_t$ to the latent at each denoising step and let the spatial frame
translate continuously along the generative trajectory, so the Euclidean
border of the latent tensor is repositioned across the $S^1$ boundary
throughout sampling and no static seam can form. After the final step, a
single inverse shift compensates the accumulated translation and realigns
the latent with the canonical conditioning, and the result is decoded with
the circular-padded VAE decoder.

\begin{figure*}[t]
    \centering
    \includegraphics[width=\linewidth]{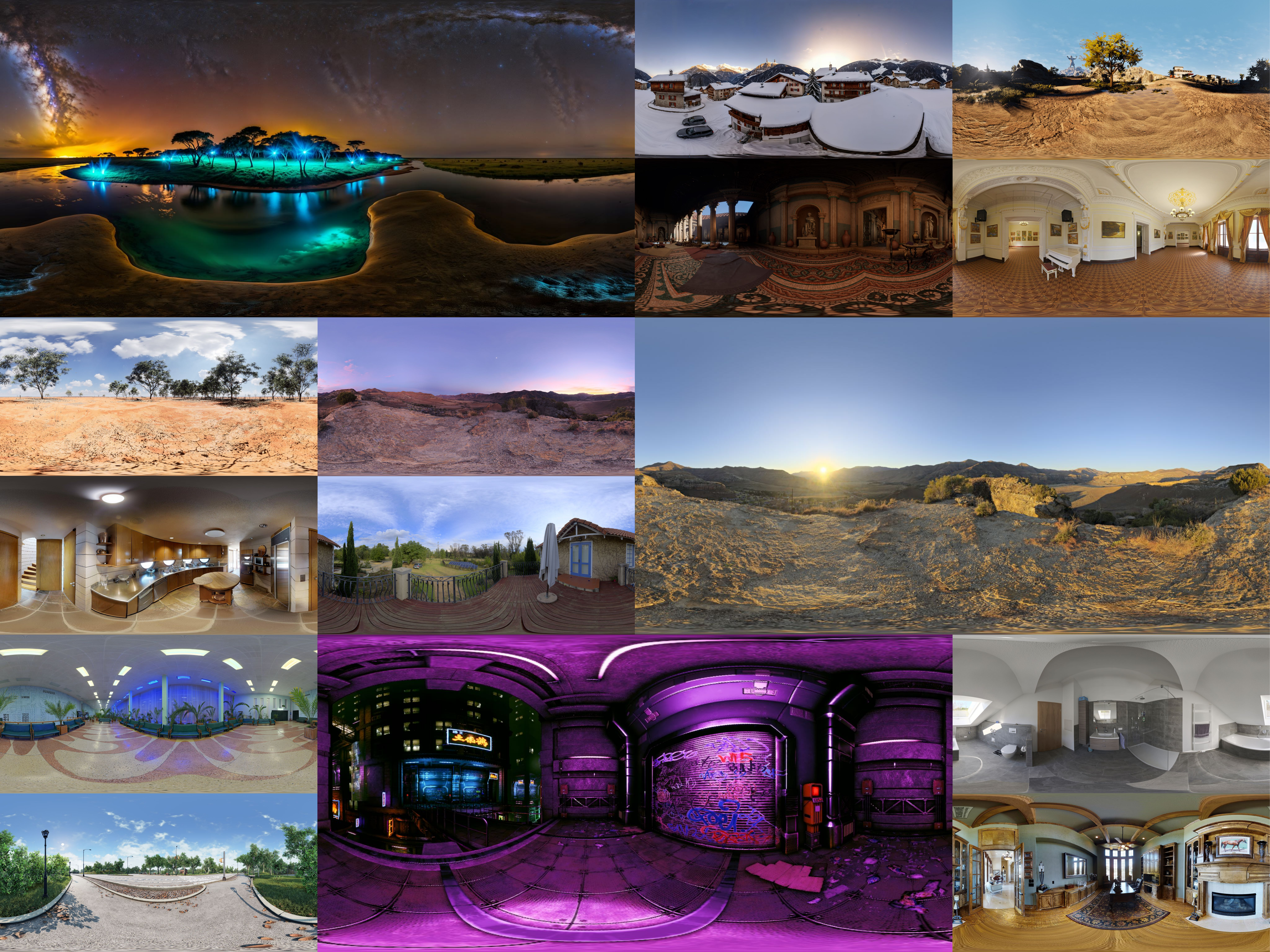}
    
    \caption{\textbf{Horizon360 Dataset Diversity.} A representative sample of our curated, gravity-aligned panoramas. The dataset spans a wide domain of indoor structures and outdoor landscapes under diverse lighting conditions, providing a robust, generalized training signal for the Gimbal360 framework.}
    \label{fig:dataset_samples}
\end{figure*}

\section{Horizon360 Dataset Details}
\label{sec:supp_dataset}

\subsection{Curation and Captioning}
Horizon360 comprises 20k high-fidelity panoramic environments aggregated
from open-source repositories and high-quality synthetic renders,
spanning complex indoor scenes with intricate lighting and dense
geometric structure as well as expansive outdoor scenarios. Text prompts are generated with a vision--language model
(e.g., Qwen3-VL \citep{Qwen3-VL}) from the canonicalized
panoramas. Fig.~\ref{fig:dataset_samples} shows a random selection of
canonicalized ground-truth panoramas, illustrating the diversity of the
dataset.

\begin{figure}[htbp]
  \centering
  \includegraphics[width=\linewidth]{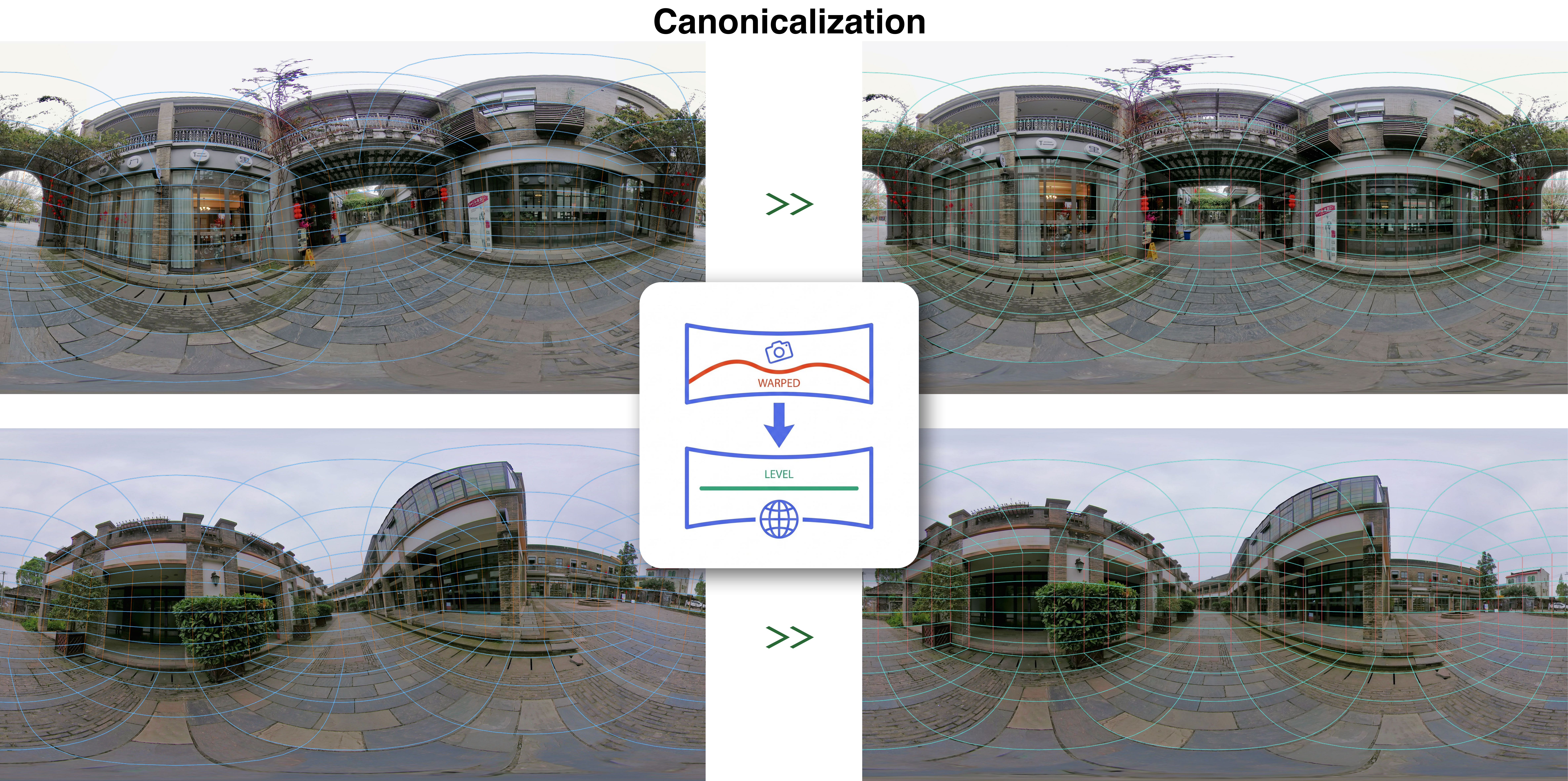}
  
  \caption{\textbf{Dataset Canonicalization.} Visualization of the rectification process applied to Horizon360. Raw unaligned panoramas (left) exhibiting severe sinusoidal distortions are mathematically rotated into the Canonical Viewing Space (right), ensuring a perfectly straight equator and plumb architectural verticals.}
  \label{fig:canonicalization}
\end{figure}

\subsection{Canonicalization}
Every panorama is rectified prior to training. As illustrated in
Fig.~\ref{fig:canonicalization}, raw unaligned panoramas exhibiting sinusoidal
distortions are rotated on the sphere so that the true horizon aligns
with the geometric equator of the ERP and vertical lines
are rendered plumb, providing the gravity-aligned structural prior
required by the Canonical Viewing Space.

\subsection{Mixture-Distribution Pose Sampling}
To improve robustness on unposed, in-the-wild photographs, we sample
perspective crops that simulate realistic captures. Real user photos mix
small hand-held jitter with occasional extreme, intentional angles, which
neither a pure Gaussian nor a pure uniform prior captures. We therefore
sample each parameter with a mixture-distribution strategy: 
\begin{itemize}
  \item \textbf{Yaw.} Sampled uniformly over $(-\pi, \pi)$ for unbiased
  azimuthal coverage.
  \item \textbf{Pitch.} $70\%$ from $\mathcal{N}(0, 15^\circ)$ simulating
  eye-level photography and $30\%$ from $\mathcal{U}(-45^\circ, 45^\circ)$
  covering extreme upward and downward shots.
  \item \textbf{Roll.} $80\%$ from $\mathcal{N}(0, 5^\circ)$ simulating
  handheld instability and $20\%$ from $\mathcal{U}(-45^\circ, 45^\circ)$
  covering aggressive tilt.
  \item \textbf{Field of view.} $80\%$ from $\mathcal{N}(60^\circ, 10^\circ)$ and $80\%$ from $\mathcal{U}(45^\circ, 100^\circ)$, covering wide-angle to telephoto optics.
\end{itemize}

\begin{figure*}[htbp]
    \centering
    \includegraphics[width=\linewidth]{fig/OOD1.pdf}
    \caption{Out-of-Distribution Generalization.}
    \label{fig:ood_generalization}
\end{figure*}

\section{Topological Equivariance}
\label{sec:supp_proof}

The main paper introduces a Siamese shift-equivariance objective for Topologically Equivariant Generation (TEG) that natively governs the diffusion process.
Here we analyze the conditions under which this objective yields
shift-equivariant denoising on the $S^1$ boundary, and then describe our
memory-efficient implementation via positional-encoding manipulation.

\subsection{Problem Formulation and $S^1$ Topology}
Let $Z \in \mathbb{R}^{C \times H \times W}$ denote the latent of an ERP
panorama, where $W$ spans the azimuth $\theta \in [-\pi, \pi]$. A true
omnidirectional environment lies on $S^2$, and its cylindrical projection onto the azimuthal axis forms a
periodic $S^1$ boundary. Mathematically, this enforces a strict boundary condition for any integer $k$:

$$Z[:, :, w] \equiv Z[:, :, (w + k \cdot W) \bmod W].$$
Standard latent diffusion architectures violate this condition through
bounded spatial priors that treat $w = 0$ and $w = W - 1$ as maximally
distant.

\subsection{Equivariance under the Siamese Objective}
We define the circular translation operator
$\mathrm{Roll}_\delta$ with
$[\mathrm{Roll}_\delta(Z)]_{:,:,w} = Z_{:,:,(w-\delta) \bmod W}$, which shifts a tensor along the azimuthal axis $W$ by an integer offset $\delta\in [0,W-1]$. 
With combined input $X = z_t \oplus M \oplus z_{\mathrm{canonical}}$, the
Siamese objective minimizes
$$  \mathcal{L}_{\mathrm{shift}} = \mathbb{E}_{X,t,\delta}
  \big[\, \lVert \mathrm{Roll}_\delta(\epsilon_\theta(X, t)) -
  \epsilon_\theta(\mathrm{Roll}_\delta(X), t) \rVert_2^2 \,\big].$$
In the limit $\mathcal{L}_{\mathrm{shift}} \to 0$, the properties of the
$L_2$ norm imply
$\mathrm{Roll}_\delta(\epsilon_\theta(X, t)) =
\epsilon_\theta(\mathrm{Roll}_\delta(X), t)$ almost everywhere, so the
denoiser's prediction cannot depend on the absolute azimuthal coordinate.
Had the network learned an artificial boundary feature at a fixed
position (e.g., a seam at $w=0$), shifting the input would move the content while the artifact
remained static, violating this equality. The objective therefore drives
the denoiser to process the latent as a closed loop along the azimuth.

Two caveats bound this analysis. It holds in the limit of a converged
Siamese loss rather than unconditionally, and it constrains the denoiser
in latent space. Since pre-trained autoencoders are not in general
equivariant under spatial transformations \citep{kouzelis2025eq},
latent-space consistency does not formally transfer to the pixel domain.
Empirically, the learned consistency yields perceptually seamless
decoded outputs, which the seam-specific metrics quantifies.

\begin{figure*}[htbp]
    \centering
    \includegraphics[width=.9\linewidth]{fig/OOD2.pdf}
    \caption{Out-of-Distribution Generalization.} 
    \label{fig:ood_generalization2}
\end{figure*}

\begin{figure*}[htbp]
    \centering
    \includegraphics[width=.9\linewidth]{fig/OOD3.pdf}
    \caption{Out-of-Distribution Generalization.} 
    \label{fig:ood_generalization3}
\end{figure*}

\subsection{Efficient Implementation via Positional Shifting}
Physically rolling high-dimensional latent concatenations in memory at
every training step introduces avoidable overhead.Because diffusion transformers process images as sequences of tokens whose spatial structure is injected purely through the positional encoding
$P(x, y)$, a circular shift of the image tensor is mathematically
identical to an inverse circular shift of the positional encodings,
$$  P_{\mathrm{shifted}}(x, y) = P((x - \delta) \bmod W,\, y).$$
During training, the base branch processes the canonical inputs with the
default coordinate grid, while the shifted branch processes the identical
unshifted tensor conditioned on the circularly shifted encodings. The
fast memory-level $\mathrm{Roll}_\delta$ is applied only to the base
branch's output to align the two predictions for the $L_2$ penalty,
avoiding duplicate memory allocation for the inputs.
At inference, the denoiser runs a single
forward pass per step, exactly as in the standard pipeline. The random
per-step shift $\delta_t$ described in Sec.~\ref{sec:supp_exp_setup} amounts
to one circular index remapping per step, applied either to the latent
directly or, equivalently, to its positional encodings, and a single
inverse shift realigns the final latent before decoding. TEG therefore
adds no additional network evaluations and negligible overhead at
inference.

\begin{figure}[htbp]
    \centering
    \includegraphics[width=\linewidth]{fig/seam.pdf}
    \caption{\textbf{Boundary Continuity Comparison.} A zoomed-in analysis of the $S^1$ azimuthal seam. Baseline method (left) relying on inference-time circular blending suffer from unnatural texture smoothing and relocated secondary seams at the blend edges. Gimbal360 (right) natively generates seamless boundaries with preserved high-frequency details and structural integrity.}
    \label{fig:seamless_boundary}
\end{figure}

\begin{figure}[htbp]
    \centering
    \includegraphics[width=\linewidth]{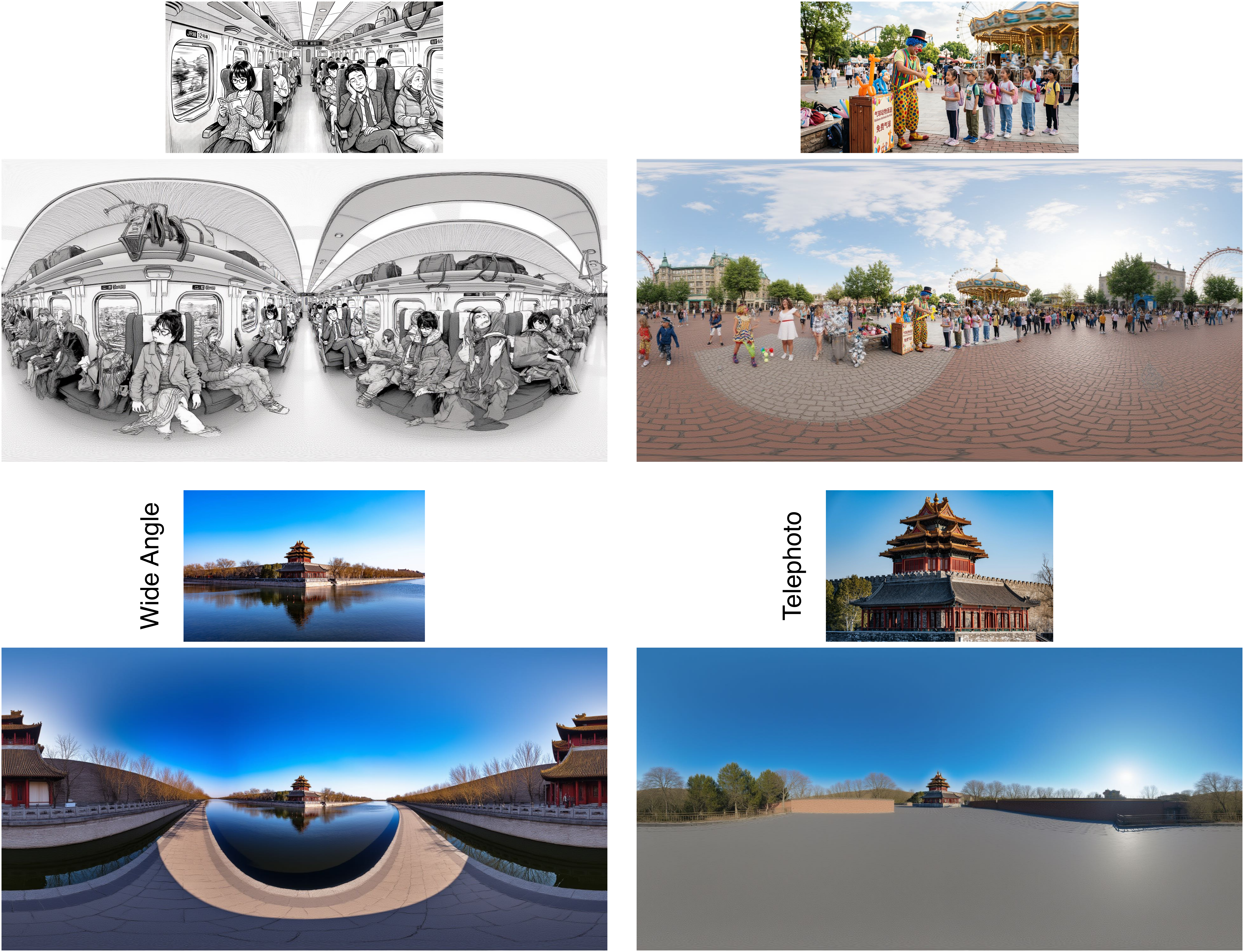}
    \caption{\textbf{Limitations and Failure Cases.} (Top) Generation with humans leads to inaccurate anatomical distortions. (Bottom) An extreme telephoto input (right) provides insufficient context, occasionally resulting in completion with less coherent global geometry than a wide-angled input (left).}
    \label{fig:limitations}
\end{figure}

\section{Additional Qualitative Results}
\label{sec:supp_qualitative}

\subsection{Out-of-Distribution Generalization}
We test Gimbal360 on challenging out-of-distribution inputs collected
from the evaluation suites of contemporaneous works
\citep{hunyuanworld2025tencent,worldgen2025ziyangxie,worldlabs}, together
with self-collected smartphone captures and AI-generated images. These
inputs feature unconstrained environments and artistic lighting that
deviate substantially from Horizon360. As shown in
Fig.~\ref{fig:ood_generalization},~\ref{fig:ood_generalization2}, and~\ref{fig:ood_generalization3}, Gimbal360 anchors these inputs into the Canonical Viewing Space and completes geometrically plausible, immersive $360^\circ$ environments.

\subsection{Seamless Synthesis at the $S^1$ Boundary}
Fig.~\ref{fig:seamless_boundary} compares zoomed-in boundary regions against
WorldGen \citep{worldgen2025ziyangxie}. Methods that rely on
inference-time circular blending exhibit two characteristic artifacts.
Averaging conflicting values across the boundary destroys high-frequency
detail, producing smoothed or ghosted textures at the seam, and the
abrupt transition at the edge of the blending zone often relocates the
seam to a new position rather than removing it. In contrast, Gimbal360
enforces continuous spatial logic during generation through TEG, yielding
perceptually seamless boundaries with preserved high-frequency detail and
structural coherence, without any destructive inference-time blending.

\subsection{Limitations and Failure Cases}
Fig.\ref{fig:limitations} illustrates two primary limitations rooted in
single-view geometry and latent diffusion priors. When the input is heavily 
dominated by human subjects, synthesizing articulated anatomy under ERP
distortion remains challenging for the diffusion prior, occasionally
producing unnatural scaling and unsatisfactory results with severely distorted figures. Additionally, extreme telephoto inputs provide little structural context to anchor a full environment, forcing the prior to hallucinate most of the canvas with minimal conditioning, which can degrade global layout
coherence and fidelity.

\end{document}